\documentclass[runningheads]{llncs}
\usepackage{graphicx}
\usepackage{tikz}
\usepackage{comment}
\usepackage{amsmath,amssymb} % define this before the line numbering.
\usepackage{color}
\usepackage{booktabs}
\usepackage{multirow}

\usepackage{paralist}

\usepackage[ruled,vlined]{algorithm2e}

\usepackage[pagebackref,breaklinks,colorlinks]{hyperref}
\usepackage{breakcites}

\usepackage[accsupp]{axessibility}  % Improves PDF readability for those with disabilities.

\usepackage{wrapfig}

\makeatletter
\newcommand{\printfnsymbol}[1]{%
  \textsuperscript{\@fnsymbol{#1}}%
}
\makeatother

\begin{document}
% \linenumbers
\pagestyle{headings}
\mainmatter
\def\ECCVSubNumber{5677}  % Insert your submission number here

\title{Dynamic Spatio-Temporal Specialization Learning for Fine-Grained Action Recognition}  % Replace with your title

\begin{comment}
\titlerunning{ECCV-22 submission ID \ECCVSubNumber} 
\authorrunning{ECCV-22 submission ID \ECCVSubNumber} 
\author{Anonymous ECCV submission}
\institute{Paper ID \ECCVSubNumber}
\end{comment}
%******************

% \begin{comment}
% \titlerunning{Dynamic Spatio-Temporal Specialization Learning}
\titlerunning{DSTS for Fine-Grained Action Recognition}
\author{Tianjiao Li\inst{1}\thanks{equal contribution} \and
Lin Geng Foo\inst{1}\printfnsymbol{1} \and
Qiuhong Ke\inst{2} \and
Hossein	Rahmani\inst{3} \and
Anran Wang\inst{4} \and
Jinghua Wang\inst{5} \and
% Jinghua Wang\inst{5} \orcidID{0000-0002-2629-1198} \and
Jun Liu\inst{1}\thanks{corresponding author}
}
\authorrunning{T. Li et al.}
\institute{ISTD Pillar, Singapore University of Technology and Design\\
\email{\{tianjiao\_li,lingeng\_foo\}@mymail.sutd.edu.sg,}
\email{jun\_liu@sutd.edu.sg}
\and
Department of Data Science \& AI, Monash University\\
\email{qiuhong.ke@monash.edu}
\and
School of Computing and Communications, Lancaster University\\
\email{h.rahmani@lancaster.ac.uk}
\and
ByteDance\\
\email{anranwang1991@gmail.com}
\and
School of Computer Science and Technology, Harbin Institute of Technology\\
\email{wangjinghua@hit.edu.cn}
}
% \end{comment}
%******************
\maketitle

%%%%%%%%% ABSTRACT
\begin{abstract}
% \vspace{-0.4em}
The goal of fine-grained action recognition is to successfully discriminate between action categories with subtle differences.
To tackle this, we derive inspiration from the human visual system which contains specialized regions in the brain that are dedicated towards handling specific tasks.
We design a novel Dynamic Spatio-Temporal Specialization (DSTS) module, which consists of specialized neurons that are only activated for a subset of samples that are highly similar.
During training, the loss forces the specialized neurons to learn discriminative fine-grained differences to distinguish between these similar samples, improving fine-grained recognition.
Moreover, a spatio-temporal specialization method further optimizes the architectures of the specialized neurons to capture either more spatial or temporal fine-grained information, to better tackle the large range of spatio-temporal variations in the videos.
% Lastly, we design an Upstream-Downstream Learning algorithm to optimize our model's dynamic decisions, allowing our DSTS module to generalize better.
Lastly, we design an Upstream-Downstream Learning algorithm to optimize our model's dynamic decisions during training, improving the performance of our DSTS module.
We obtain state-of-the-art performance on two widely-used fine-grained action recognition datasets. 
% We will release our code.
%\ke{too long abstract?} \lgg{is this length short enough?}
\end{abstract}

%   \lgg{Note: focus more on the spatio-temporal aspect}

%   \lgg{Note:Make different from other work.}

% \vspace{-1.1em}
\section{Introduction}
% \vspace{-0.3em}

Fine-grained action recognition involves distinguishing between similar actions with only subtle differences, e.g., ``cutting an apple in a kitchen" and ``cutting a pear in a kitchen".
This is significantly more challenging than coarse-grained classification, where the action classes can be ``cutting something in a kitchen" and ``playing in a gym". The higher inter-class similarity 
in the fine-grained setting makes it a challenging task, which 
coarse-grained backbones and methods 
% \cite{feichtenhofer2019slowfast,lin2019tsm,yang2020tpn,carreira2017i3d,wang2016tsn} 
struggle to overcome.

To tackle the challenging fine-grained
action recognition task,
we derive inspiration from
the remarkable human visual system which has good
fine-grained recognition capabilities.
Importantly, our visual system comprises of specialized neurons that are activated only under some specific circumstances, as shown by previous works \cite{tsao2008comparing,minxha2017}.
For example, for enhanced recognition of humans 
which is crucial for social behaviour, human brains have developed a set of cortical regions specialized for processing faces \cite{tsao2008comparing,minxha2017}. These specialized regions fire only when our attention focuses on human faces, while specific sub-regions are further specialized to fire specifically for processing face parts \cite{pitcher2007tms}, 
eye gazes and expressions
\cite{hoffman2000distinct}, and identity \cite{rotshtein2005morphing,haxby2000distributed}.

\begin{wrapfigure}[25]{r}{0.5\linewidth}
% \begin{wrapfigure}{r}{0.5\linewidth}
    \centering
    \includegraphics[width=\linewidth]{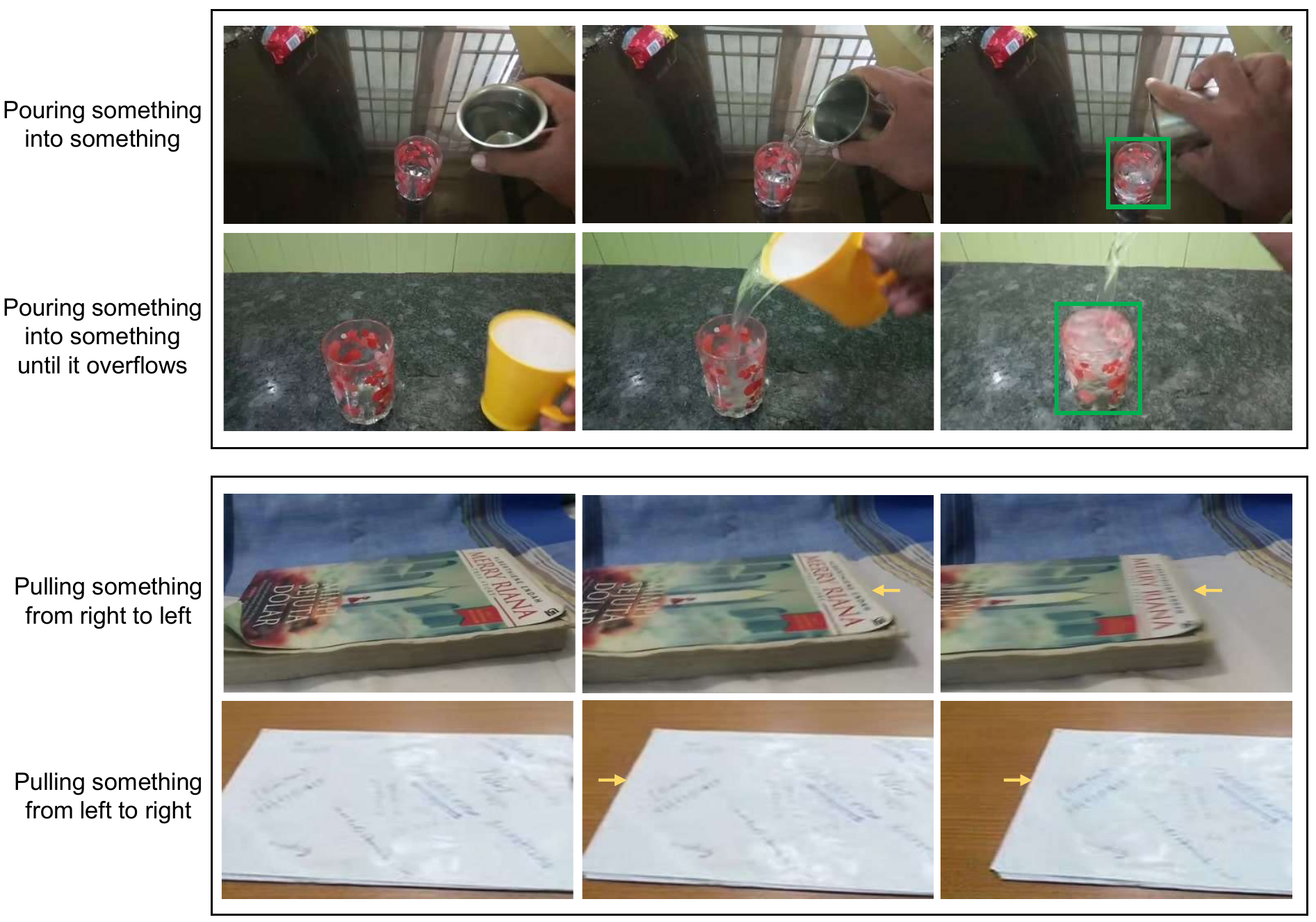}
    \caption{Key frames of samples taken from the Something-Something-v2 dataset \cite{goyal2017something}. (Top) Fine-grained differences lie more in the spatial aspects of the two actions, as shown in the green box. To distinguish between these two actions, we need to focus on whether the water in the cup overflows in the final key frame, which can be quite subtle. (Bottom) Fine-grained differences lie mainly in the temporal aspects of the two actions, where %the focus is 
    we need to focus on the movement (denoted with yellow arrows) of the object across all key frames.
    Best viewed in colour.
    }
    \label{fig:examples}
\end{wrapfigure}

Inspired by the specialization of neurons in the human brain, we
improve fine-grained recognition capabilities of a neural network by using specialized parameters 
that are only activated 
on a subset of the data.
More specifically, we design a novel Dynamic Spatio-Temporal Specialization (DSTS) module which consists of \textit{specialized neurons} that are only activated when the input is within their area of specialization (as determined by their individual \textit{scoring kernels}). 
In particular, a synapse mechanism dynamically activates each specialized neuron only on a subset of samples that are highly similar, such that only fine-grained differences exist between them.
During training, in order to distinguish among that particular subset of similar samples, the loss will push the specialized neurons to focus on exploiting the fine-grained differences between them.
We note that previous works on fine-grained action recognition \cite{zhang2021tqn,wu2019longterm,zhou2015interaction} have not explicitly considered such specialization of parameters. These works \cite{zhang2021tqn,wu2019longterm,zhou2015interaction} propose deep networks where all parameters are generally updated using all data samples, and thus, during training, the loss tends to encourage those models to pick up 
more common discriminative cues
that apply to the more common samples,
as opposed to various fine-grained cues that might be crucial to different subsets of the data.

Another interesting insight comes from the human primary visual cortex, where there are neurons that are observed to be specialized in temporal or spatial aspects \cite{xu2001comparison,nolte2016nolte}. Magnocellular, or M cells, are observed to be specialized to detect movement, e.g., speed and direction. Parvocellular, or P cells, are important for spatial resolution, e.g., shape, size and color. Together, they effectively allow humans to distinguish between actions.

Spatial and temporal specialization has clear benefits for fine-grained action recognition.
As observed in Fig \ref{fig:examples}, some fine-grained differences lie mainly in the temporal %aspect of an action, 
aspects of two actions,
e.g., %between 
``Pulling something from right to left" and ``Pulling something from left to right". In this case, a greater emphasis on the temporal dimension of each video %dimensions of features 
will lead to better recognition performance.
In contrast,
some fine-grained differences lie more in the spatial aspects of two actions, %aspect of the action,
e.g., ``Pouring something into something" and ``Pouring something into something and it overflows". 
In this case, greater emphasis on the spatial dimension can improve the performance.
%When crucial fine-grained differences exist in the temporal aspect of an action, a greater focus on the temporal dimensions of features  will lead to better  performance. On the other hand, when fine-grained differences exist in the spatial aspect %of an action, additional focus on the spatial dimensions will improve the performance. 

% \lgg{We need diversified experts}
% How can we obtain a more diverse set of specialized neurons with different specializations that can better tackle a wider variety of fine-grained differences?
% Moreover, it benefits our DSTS module to have a set of specialized neurons with highly diversified specializations, so that the module can handle a more diverse range of fine-grained differences.

To allow our module to efficiently and effectively handle fine-grained differences over a large range of spatio-temporal variations, we design a \textit{spatio-temporal specialization} method that 
additionally provides specialized neurons with spatial or temporal specializations. To achieve such specialization, we explicitly design the specialized neurons to focus only on one single aspect (spatial or temporal) for each  channel of the input feature map at a time, forcing the neurons to exploit fine-grained differences between similar samples in that specific aspect,  %only,
leading to higher sensitivity towards these fine-grained differences. 
Specifically, this is implemented using \textit{gates} that determine whether a \textit{spatial operator} or a \textit{temporal operator}
is used to process each input channel. 
By adjusting their gate parameters, specialized neurons that benefit from discerning spatial or temporal patterns adapt their architectures 
to use the corresponding operator across more channels.
Eventually, the set of specialized neurons will have diversified architectures and specializations focusing on different spatial and temporal aspects, which collectively are capable of handling a large 
variety of spatial and temporal fine-grained differences.

% To optimize our DSTS module to better generalize to unseen samples, we design an Upstream-Downstream Learning algorithm that optimizes the model parameters involved in making dynamic decisions,
% which we call \textit{upstream parameters}.
% These upstream parameters (i.e., scoring kernels and gate parameters) have profound impacts on the training of \textit{downstream parameters} (i.e., spatial and temporal operators). We improve generalization capability by optimizing
% upstream parameters  while taking their downstream effects 
% on the spatial and temporal operators
% into account. 

% \textcolor{red}{
% To further improve the performance of our DSTS module, we design an Upstream-Downstream Learning (UDL) algorithm that better optimizes the model parameters involved in making dynamic decisions, which we call \textit{upstream parameters}.
% During end-to-end training of our model, we jointly train
% these upstream parameters (i.e., scoring kernels and gate parameters) that make dynamic decisions and \textit{downstream parameters} (i.e., spatial and temporal operators) that process input, which can be challenging as \textit{upstream parameters themselves also affect the training of downstream ones}.
% We thus improve performance by optimizing upstream parameters through meta-learning, such that they learn how to make decisions that positively affect the training of downstream parameters.
% }

During end-to-end training of our module, we jointly train two types of parameters:
\textit{upstream parameters} (i.e., scoring kernels and gate parameters) that make dynamic decisions and \textit{downstream parameters} (i.e., spatial and temporal operators) that process input, which can be challenging as \textit{upstream parameters themselves also affect the training of downstream ones}.
Hence, we design an Upstream-Downstream Learning (UDL) algorithm to optimize upstream parameters to learn how to make decisions that positively affect the training of downstream parameters, improving the performance of our DSTS module.
% through the use of meta-learning.

% % %%%%%%%%% comment out in camera-ready, for space %%%%%%%%%
% In summary, our contributions are: \textbf{1)} Inspired by the specialization of neurons in the human brain, we propose our novel DSTS module to tackle fine-grained action recognition. DSTS is a dynamic module which only activates specialized neurons with the relevant fine-grained specialization to process each sample for better fine-grained recognition. \textbf{2)} To allow our DSTS module to efficiently and effectively handle a large variety of spatial and temporal fine-grained differences, we further optimize the architectures of specialized neurons (towards spatial or temporal fine-grained aspects) with a spatio-temporal specialization method. 
% \textbf{3)} To improve the performance of our DSTS module, we design a UDL procedure.
% \textbf{4)} We obtain improved performance after adding DSTS to the baseline CNN and Transformer backbones, surpassing the state-of-the-arts on two popular fine-grained action recognition datasets.

\section{Related Work}
\subsection{Action Recognition}
Action recognition involves taking an action video clip as input and predicting the class of the action.
% Many methods have been proposed to tackle this task, including 2D networks such as the two-stream \cite{simonyan2014twostream}, TSN \cite{wang2016tsn},
% TRN \cite{zhou2018trn}, TSM \cite{lin2019tsm}, TPN \cite{yang2020tpn}, 3D networks such as
% LTC \cite{varol2017ltc}, I3D \cite{carreira2017i3d}, S3D \cite{xie2018s3d}, SlowFast \cite{feichtenhofer2019slowfast}, X3D \cite{feichtenhofer2020x3d} and NL \cite{wang2018nonlocal}. 
Many methods have been proposed to tackle this task, including the two-stream \cite{simonyan2014twostream}, TSN \cite{wang2016tsn},
TRN \cite{zhou2018trn}, TSM \cite{lin2019tsm}, TPN \cite{yang2020tpn},
LTC \cite{varol2017ltc}, I3D \cite{carreira2017i3d}, S3D \cite{xie2018s3d}, SlowFast \cite{feichtenhofer2019slowfast}, X3D \cite{feichtenhofer2020x3d}, NL \cite{wang2018nonlocal}, GST \cite{luo2019grouped}, Tx \cite{girdhar2019actiontransformer}, TimeSformer \cite{bertasius2021space}, ViViT \cite{arnab2021vivit}, MViT-B \cite{fan2021multiscale}, and Swin Transformer \cite{liu2021video}. 
% Unlike these works, we focus on the fine-grained setting.

%and various mixed-transformer architectures such as Tx \cite{girdhar2019actiontransformer} and NL \cite{wang2018nonlocal}. 

 %LFB \cite{wu2019longterm}, TQN \cite{zhang2021tqn}.
% GST \cite{luo2019grouped}.

\subsection{Fine-grained Action Recognition}
In comparison, fine-grained action recognition,
where actions have lower inter-class differences, has been relatively less explored. Datasets such as Something-Something-v2 \cite{goyal2017something} and Diving48 \cite{li2018diving48}
% and FineGym \cite{shao2020finegym} 
have been curated for this purpose.
% Methods that directly tackle the fine-grained task includes 

Interaction Part Mining \cite{zhou2015interaction} mines mid-level parts, connects them to form a large spatio-temporal graph and mines interactions within the graph. LFB \cite{wu2019longterm} employs a long-term feature bank for detailed processing of long videos that provides video-level contextual information at every time step. 
% Attention-guided LSTM \cite{kanojia2019attentive} uses an attention network to focus attention on the important spatial locations in the video. 
% CorrNet %Video Modeling with Correlation Networks
FineGym \cite{shao2020finegym} has found 
that coarse-grained backbones lack the capability to capture complex temporal dynamics and subtle spatial semantics for their fine-grained dataset.
TQN \cite{zhang2021tqn} casts fine-grained action recognition as a query-response task, where the model learns query vectors that are decoded into response vectors by a Transformer.

Different from previous works that do not explicitly consider specialized parameters, we propose a novel dynamic DSTS module that trains and selects specialized neurons %with the most relevant fine-grained specialization 
for fine-grained action recognition. Furthermore, we investigate a novel spatio-temporal specialization scheme that optimizes architectures of the specialized neurons to focus more on spatial or temporal aspects, further specializing them for improved fine-grained action recognition. 
% A meta-learning method is also proposed for better generalizability.
%Furthermore, our specialized neurons also optimize their architectures to focus more on spatial or temporal aspects, to further specialize in discriminating between actions that have fine-grained spatial or temporal differences.
% Furthermore, our specialized neurons learn to focus more on spatial or temporal aspects, to specialize in discriminating between actions that have fine-grained spatial or temporal differences.
% Our approach to solving the fine-grained action recognition problem is different from the above works, as we attempt to tackle the underlying issues of ``lazy" behaviour and noise from non-localized gradients through the dynamic use of convolutional filters.

% \begin{itemize}
%     \item Attention mechanism/bounding box
%     \item Higher order representation learning
% \end{itemize}
%maybe delete ha2016hypernetworks,jia2016dynamic and 
%okay

%% ha2016hypernetworks,jia2016dynamic
%%mullapudi2018hydranets,shazeer2017outrageously,
\subsection{Dynamic Neural Networks} 
% Dynamic neural networks generally adapt their parameters or structures according to the input. Typical approaches include generating weights with a subnetwork \cite{zamora2019adaptive,tian2020conditional}, dynamically selecting network depth \cite{wang2018skipnet,veit2018convaig} and dynamically selecting network widths \cite{hua2019channel,gao2019dynamic}.

Dynamic neural networks generally adapt their parameters or structures according to the input. Typical approaches include generating weights with a subnetwork, dynamically selecting network depth and dynamically selecting network widths 
% \cite{zamora2019adaptive,wang2018skipnet,hua2019channel}. 
\cite{zamora2019adaptive,wang2018skipnet,hua2019channel,wu2018blockdrop,zhang2021learning}. 
% For segmentation purposes, some methods explore usage of a dynamic network to generate weights to tackle specific spatial locations \cite{xie2020spatially,almahairi2016dynamic,wang2019elastic}. 
On videos, 
% Adaframe \cite{wu2019adaframe} adaptively selects the frames to observe next. 
% AR-Net \cite{meng2020ar} adaptively selects the resolution of each frame to reduce computational cost. 
several methods \cite{wu2019adaframe, wu2020dynamic} adaptively select video frames for the sake of efficiency.
GSM \cite{sudhakaran2020gate} learns to adaptively route features from a 2D-CNN through time and combine them.
TANet \cite{liu2021tam} employs a dynamic video aggregation kernel that adds global video information to 2D convolutions. 
% SGS \cite{fayyaz20213d} dynamically altered the temporal resolution of features to reduce computational cost. 
% \lgg{maybe can find more...}
% Several methods adaptively select video frames \cite{wu2019adaframe, wu2020dynamic} for the sake of efficiency.
% \lgg{Cite the ERA paper?}

% Our DSTS module dynamically selects the most suitable specialized neurons, and is a type of dynamic architecture. 
Different from these methods, our DSTS module focuses on  improving performance on fine-grained action recognition.
We design a novel synapse mechanism that activates each specialized neuron only on samples that are highly similar, pushing them to pick up relevant fine-grained differences to distinguish between these similar samples.
% Different from previous methods, our DSTS module focuses on dynamically activating each specialized neuron on samples that are highly similar, pushing them to pick up relevant fine-grained details to distinguish between these similar samples and improving performance on fine-grained action recognition.
We further propose spatio-temporal specialization of our specialized neurons, which to the best of our knowledge, has not yet been explored in previous works.
% We also investigate a novel spatio-temporal specialization that further specializes the architecture of specialized neurons to handle spatial or temporal fine-grained aspects.
% widening the diversity of specializations within a DSTS module to cover a wider variety of fine-grained differences.
% capabilities on the fine-grained action recognition task.

\subsection{Kernel Factorization} 
Kernel Factorization generally involves factorizing a 3D spatio-temporal convolution into a 2D spatial convolution plus a 1D temporal convolution, such as in P3D \cite{qiu2017learning}, S3D \cite{xie2018s3d} and R(2+1)D \cite{tran2018closer}.
% , where kernel factorization led to improved effectiveness and efficiency. 
In GST \cite{luo2019grouped}, 3D convolutions are decomposed into a fixed combination of parallel spatial and temporal convolutions.
% fixed combination of spatial and temporal convolutions that are conducted in parallel.
In these works, the kernel factorization leads to improved effectiveness and efficiency. 

% Here, we propose a novel DSTS module that automatically and dynamically xxx, for handling the fine-grained differences in fine-grained action recognition.
Here, we propose a novel DSTS module that dynamically activates the most relevant specialized neuron.
Different from previous works, our specialized neurons learn to select a spatial or temporal operator for each channel, to better handle the 
% spatial and temporal 
corresponding
fine-grained differences between similar samples, for fine-grained action recognition.

\section{Proposed Method}

%different layers? number of layers?
%skip connection
%number of specialized neurons
% spatial output Z_S and temporal output Z_T
%n_best

\begin{wrapfigure}{r}{0.5\linewidth}
    % \vspace{-0.6cm}
    \centering
    \includegraphics[width=\linewidth]{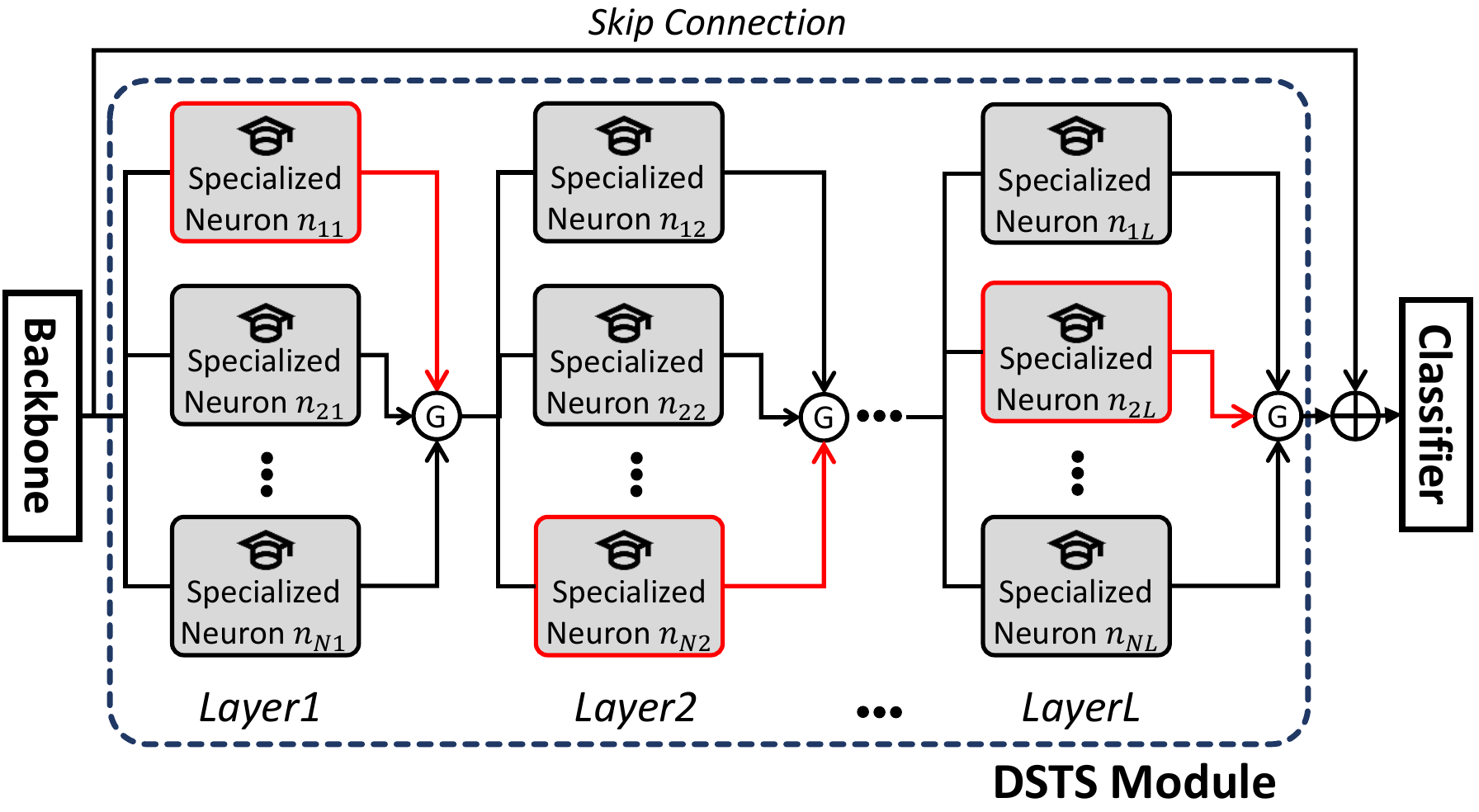}
    % \vspace{-0.4cm}
    \caption{ Illustration of the proposed DSTS module, which processes features extracted from a backbone.
    There are $L$ layers within the DSTS module, each comprising $N$ specialized neurons (grey rectangles). 
    When a feature map $X$ is 
    fed into the $j$-th DSTS layer, impulse values $v_{ij}$ from each specialized neuron $n_{ij}$ are first calculated, and the specialized neuron with the highest impulse value in that layer is activated (indicated with red arrows) using the Gumbel-Softmax technique (indicated with 
    % \raisebox{.5pt}{\textcircled{\raisebox{-.9pt} {G}}}
    {\large \textcircled{\small G}}).
    A skip connection adds general features from the backbone to the output of the DSTS module (indicated with $\bigoplus$), before being fed into the classifier.
    }
    \label{fig:DSTS}
% \vspace{-0.36em}
\end{wrapfigure}

\subsection{Overview}

In fine-grained action recognition, actions from different classes can be highly similar, 
with only fine-grained differences between them. 
Such fine-grained differences might not be effectively learnt by parameters that are trained on all samples, as they will tend to capture common discriminative cues
% general discriminative patterns %patterns and cues 
that occur more commonly throughout the data, instead of various fine-grained %information that 
cues, each of which might only be relevant in a 
small subset
of the data \cite{johnson2019survey}. Thus, to improve performance on fine-grained action recognition, we propose to employ specialized parameters in our model. These %; these
specialized parameters are pushed to gain specialized capabilities in identifying fine-grained differences by being trained only on a subset of the data that contains highly similar samples.

Our DSTS module achieves this specialization through the dynamic usage of blocks of parameters called \textit{specialized neurons}, which can be observed in Fig.~\ref{fig:DSTS}.
For each input sample, only one specialized neuron (i.e., the neuron with the most relevant specialization) in each layer is activated to process the sample -- this dynamic activation occurs in what we call the \textit{synapse mechanism}. 
Crucially, we design the synapse mechanism such that each specialized neuron is only activated on a subset of samples that are similar, with only fine-grained differences between them.
During training, since each specialized neuron is only trained on a subset of the data that contains similar samples, the training loss will push the specialized neuron to learn to handle the fine-grained information relevant to these samples, instead of learning more common discriminative 
cues that are applicable to the more % of the
common samples.
Hence, each specialized neuron gains specialized capability that is highly effective at classifying a particular subset of samples, leading to improved fine-grained recognition performance.
%\lgg{should I add a line break here?yes }

% \textcolor{red}{line break inserted}
Moreover, considering that fine-grained differences between similar samples might exist in 
more spatial or temporal aspects, 
we further propose spatio-temporal specialization in the specialized neurons, 
to further optimize their architectures.
By explicitly forcing the specialized neurons to focus on
spatial or temporal aspects for each channel of the input feature map, 
they are pushed to exploit fine-grained differences 
in that specific aspect,
leading to better sensitivity towards the fine-grained differences in that aspect.
Such channel-wise decisions on spatial or temporal specializations are learned in an end-to-end manner for improved performance. 
Lastly, we further improve the generalization capability of our DSTS module by proposing \textit{Upstream-Downstream Learning}, where the model parameters involved in making its dynamic decisions are meta-learned.

% \textcolor{red}{Several alternative designs to our proposed architecture are considered and evaluated in the Supplementary.}

\begin{wrapfigure}{r}{0.5\linewidth}
    \centering
    \includegraphics[width=0.95\linewidth]{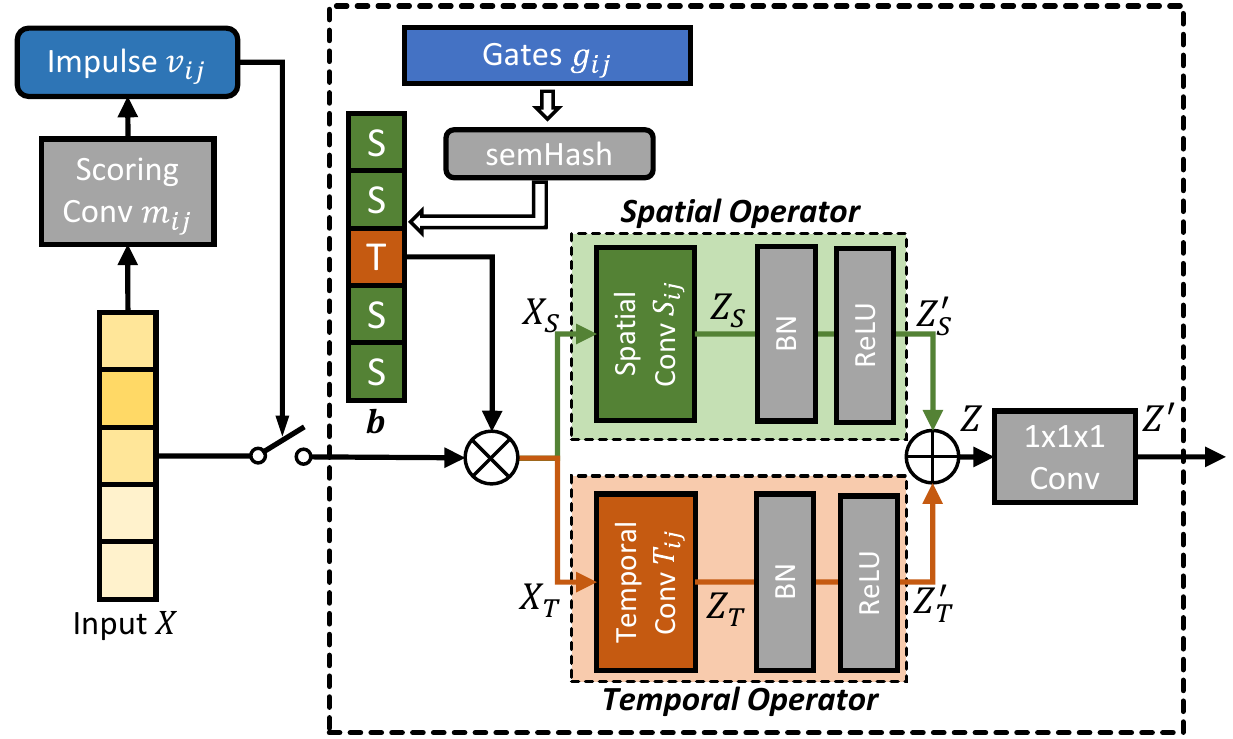}
    \caption{
    Illustration of a specialized neuron $n_{ij}$.
    % after spatio-temporal specialization. 
    % A scoring convolution $m_i$ on input $X$ produces an impulse score that determines activation within the synapse mechanism.
    A scoring convolution using $m_{ij}$, followed by a summation, produces impulse $v_{ij}$ that is used to determine if $n_{ij}$ is activated.
    Gate parameters $g_{ij}$ are used to generate $\textbf{b}$ using the Improved Semhash method, which determines (using channel-wise multiplication $\bigotimes$) if each input channel uses the spatial operator's kernel 
    $S_{ij}$ 
    (in green) or the temporal operator's kernel 
    $T_{ij}$ 
    (in orange).
    After the processing of the spatial and temporal operators, both features $Z'_S$ and $Z'_T$ are added (indicated with $\bigoplus$) and fused using a $1 \times 1 \times 1$ convolution to get output $Z'$.
    % \lgg{to edit to be clearer}
    % \textcolor{red}{image edited to show interactions better}
    }
    \label{fig:st_specialization}
\end{wrapfigure}

Next, we formally introduce the DSTS module which is illustrated in Fig.~\ref{fig:DSTS}. 
Setting batch size to $1$ for simplicity, we assume that the pre-trained backbone outputs a feature map $X \in \mathbb{R}^{N_{in} \times N_t \times N_h \times N_w}$, where $N_{in},N_{t}, N_h, N_w$ represent the channel, temporal, height and width dimensions of the feature map, respectively.
The DSTS module consists of $L$ layers, with each layer comprising of $N$ specialized neurons. 
We define the $i$-th specialized neuron in the $j$-th layer as $n_{ij}$, which is shown in detail in Fig.~\ref{fig:st_specialization}.
% We illustrate a specialized neuron $n_{ij}$ in detail in Fig.~\ref{fig:st_specialization}.
Each specialized neuron $n_{ij}$ has a 
% dendrite kernel $d_i \in \mathbb{R}^{k_{t}^D \times k_{h}^D \times k_{w}^D}$, a 
scoring kernel $m_{ij} \in \mathbb{R}^{N_{out} \times N_{in} \times 1 \times 1 \times 1}$ (with the size of $1\times1\times1$ for efficiently encoding
% ``similarity"
information from all channels of feature map $X$),
% (\ke{a bit confuse of why 1x1x1 and the $n_i$ as output channel.} \lgg{this is just what tianjiao implemented >.< is there a better way to put this?}), \ke{any intuition? how this 1x1x1 match the pattern of input? } \lgg{since we set $N_{out} = N_{in}$, I have changed it to use $N_{out}$ for more clarity. I think it's mainly for efficiency and while gathering ``similarity" information from all channels. Should I write it this way? \ke{yes, maybe in the method part of the scoring? or maybe just here}} \lgg{okay.}
a spatial operator consisting of a  convolutional kernel $S_{ij} \in \mathbb{R}^{N_{out} \times N_{in} \times 1 \times 3 \times 3}$ (2D on the spatial domain),
a temporal operator consisting of a convolutional kernel $T_{ij} \in \mathbb{R}^{N_{out} \times N_{in} \times 3 \times 1 \times 1}$ (1D on the temporal domain)
and gates $g_{ij} \in \mathbb{R}^{ N_{in}}$.
% gates $g_{ij} \in \mathbb{R}^{ N_{in}}$
% \lgg{and a $1 \times 1 \times 1$ convolution kernel of shape $N_{out} \times N_{out} \times 1 \times 1 \times 1$.}

% We factorize the specialized operator (3D convolution) into a spatial operator $S \in \mathbb{R}^{N_{out} \times N_{in} \times 1 \times 3 \times 3}$ which is a 2D convolution, and the temporal operator $T \in \mathbb{R}^{N_{out} \times N_{in} \times 3 \times 1 \times 1}$ which is a 1D convolution. 

\subsection{DSTS Layer}
\label{sec:DSTS}

In this subsection, we describe a single DSTS layer. 
For clarity, we describe the first DSTS layer 
and omit the layer index, 
using $n_i$ to represent the $i$-th specialized neuron (which consists of $m_i,S_i,T_i$ and $g_i$) in this DSTS layer. 
% Several alternative designs to our proposed DSTS layer (as described in this section) are discussed in the Supplementary.

% \textcolor{red}{Several alternative designs to our proposed architecture are considered and evaluated in the Supplementary.}

% Assume that $N_{out} = N_{in}$ for ease of explanation, although it does not need to be the case.

% $Q_t = \lfloor (N_t - k_t^D + p_t^D + s_t^D)/s_t^D \rfloor, Q_h = \lfloor (N_h - k_h^D + p_h^D + s_h^D)/s_h^D \rfloor, Q_w = \lfloor (N_w - k_w^D + p_w^D + s_w^D)/s_w^D \rfloor$

% \vspace{-0.4em}

\subsubsection{Synapse Mechanism}
\label{sec:synapse}

The synapse mechanism is the crucial step that dynamically activates the specialized neuron with the most relevant specialization for the given input feature map $X$.
Importantly, similar feature maps should activate the same specialized neurons, so that each specialized neuron is pushed to specialize in fine-grained differences to distinguish between these similar feature maps during training.
% \textcolor{red}{Importantly, similar feature maps should activate the same specialized neurons, so that specialized neurons are pushed to specialize in fine-grained differences to distinguish between these similar feature maps.}

To implement the synapse mechanism to achieve the above-mentioned specialization effect, we include a \textit{scoring kernel} $m_i$  in each specialized neuron $n_i$  
% \ke{use $m_{ij}$ and $n_{ij}$ as before to be consistent and also the figure symbol?}
that is applied on the input feature map $X$ 
% \ke{X is only output by the backbone, but the intermediate layers use the output of previous layer, not  X?}
in a step that we call the \textit{scoring convolution}. The resulting output is summed to produce a 
relevance score (which we call an \textit{impulse} $v_i$)  between the input feature map $X$ and the fine-grained specialization capabilities of the specialized neuron $n_i$.
% score (which we call an \textit{impulse}) 
% that determines how relevant the input feature is, to the fine-grained specialization capabilities of the specialized neuron.
% that determines the degree of relevance of the input feature with respect to the fine-grained specialization capabilities of the specialized neuron.
% The higher the impulse of the specialized neuron is, the more relevant it is, and the more likely it will be activated.
The higher the impulse produced by a specialized neuron, the higher the relevance of the specialized neuron's knowledge to the input feature, and the more likely it will be activated. 
% As similar samples tend to have similar features, they will tend to have high impulse scores for (and activate) the same specialized neuron with the relevant knowledge.
% This achieves the aim of the synapse mechanism where specialized neurons are mostly activated by a subset of the data samples that are very similar.

% The synapse mechanism is the crucial step that dynamically activates the specialized neuron with the most relevant specialization for the given input feature map $X$.
% Importantly, similar feature maps should activate the same specialized neurons, so that each specialized neuron is pushed to discriminate between similar feature maps (with only fine-grained differences).
In the first step, to calculate the relevance scores between a specialized neuron $n_i$ and a feature map $X$, 
% we first apply its dendrite $d_i$ on $X$ in a convolutional manner. 
we first apply a scoring convolution using the scoring kernel $m_i$ on $X$:
% Using the same abuse of notation for scoring convolution function $m_i (\cdot)$ using scoring kernel $m_i$:
% \vspace{-0.3em}
\begin{equation}
    \label{eqn:scoring_conv}
    q_i = m_i (X),
    % \vspace{-0.3em}
\end{equation}
% \vspace{-0.2em}
where we slightly abuse the notation to let $m_i (\cdot)$ denote the scoring convolution function applied to %some input
an input using scoring kernel $m_i$  
(we adopt this notation for all convolution functions in this work) and
$q_i \in \mathbb{R}^{N_{out} \times Q_t \times Q_h \times Q_w}$ is an intermediate representation with $Q_t,Q_h,Q_w$ being the resulting temporal, width and height dimensions.

We then sum all elements in $q_i$ to get the impulse $v_i$ of specialized neuron $n_i$.
% \vspace{-0.5em}
\begin{equation}
    \label{eqn:impulse_sum}
    v_i = \sum_{u_c=1}^{N_{out}} \sum_{u_t=1}^{Q_t} \sum_{u_h=1}^{Q_h} \sum_{u_w=1}^{Q_w} q_{i,u_c,u_t,u_h,u_w}
\end{equation}
We conduct the above process (Eq.~\ref{eqn:scoring_conv} and Eq.~\ref{eqn:impulse_sum}) for all 
scoring kernels $\{ m_i \}_{i=1}^N$ of the $N$ specialized  neurons in the DSTS layer to obtain the complete set of impulse values $\mathcal{V}$:

% \vspace{-0.5em}
\begin{equation}
    \mathcal{V} = \{v_i\}_{i=1}^N.
\end{equation}
% As convolutional filters tend to produce highest levels of activation on suitable feature maps that share many similarities \cite{zeiler2014visualizing,Yosinski2015UnderstandingNN}, $q_i$ and $v_i$ will be similar for similar feature maps.
% As convolutional filters tend to produce highest levels of activation on suitable feature maps \cite{zeiler2014visualizing,Yosinski2015UnderstandingNN}, $q_i$ and $v_i$ will be similar for similar feature maps.

% \lgg{talk more straightforward about Gumbel softmax}

Finally, we apply the Gumbel-Softmax technique \cite{jang2016categorical} on $\mathcal{V}$ to select a specialized neuron to activate. 
The selection to activate specialized neuron $n_{a}$ is made by producing a one-hot vector with a $1$ at the selected index $a$. During training, the Gumbel-Softmax allows gradients to backpropagate through this selection mechanism.
During testing, the activated specialized neuron $n_{a}$ is the one with the highest impulse within $\mathcal{V}$,
and has the most relevant specialization to discriminate %using fine-grained information in $X$.
between samples similar to the input $X$.

We remark that this synapse mechanism is crucial for the specialization of the specialized neurons.
As convolutional filters tend to produce similar responses for similar feature maps \cite{zeiler2014visualizing,Yosinski2015UnderstandingNN}, $q_i$ and $v_i$ tend to be similar for similar feature maps. % features. 
% Hence, similar feature maps are highly likely to produce high impulse scores for (and activate) the same specialized neuron, 
% which pushes this neuron to specialize in fine-grained differences to distinguish between these features. 
Hence, during training, similar feature maps 
are highly likely to produce high impulse scores for (and activate) the same specialized neuron;
this neuron will thus be updated using only a subset of similar samples, which pushes this neuron to specialize in fine-grained differences to distinguish between them.

% \vspace{-1em}
\subsubsection{Spatio-Temporal Specialization}

Intuitively, after $n_a$ is activated, we can simply apply a  3D convolution kernel (corresponding to $n_a$) on $X$ to extract the spatio-temporal information. 
Yet, in fine-grained action recognition, the fine-grained differences between actions can exist in more spatial or temporal aspects of actions, which require emphasis along their respective dimensions for effective discrimination. 
Motivated by this, instead of 
optimizing the parameters within a 3D kernel architecture, we additionally optimize the architectures of the specialized neurons to specialize in focusing on either more spatial or more temporal fine-grained information.

More concretely, our spatio-temporal specialization method adapts the architectures of the specialized neurons to utilize either a \textit{spatial operator} or a \textit{temporal operator} for each input channel.
The spatial operator uses a 2D convolution that focuses on the spatial aspects of the feature map while the temporal operator uses a 1D convolution that focuses on the temporal aspects.
To achieve spatial or temporal specialization, we explicitly restrict the specialized neurons to choose between %(spatial or temporal)
spatial or temporal operators for each input channel.
% , forcing them to focus on only a single aspect.
During training, this design forces each specialized neuron to exploit fine-grained differences in each channel %(between similar samples) 
between similar samples in the chosen aspect, leading to better sensitivity towards these fine-grained differences.
Since different channels of the input feature map can convey different information, which might lie in the spatial or temporal aspects, we let our model adapt
its architecture to select the operator in each channel that would lead to greater discriminative capability.
Such architectural decisions (spatial or temporal) for each channel are learned by the \textit{gate} parameters.
When it is beneficial for the specialized neuron
to focus more on a certain fine-grained aspect, the gates will learn to use the corresponding operator across more channels, pushing for higher sensitivity towards that aspect for improved discriminative capability.
The efficacy of this channel-wise design for spatio-temporal specialization is investigated empirically along with other baselines in Section~\ref{sec:ablation}.

% \textbf{Spatio-temporal Architectural Decisions using Gates.} 

\noindent\underline{Spatio-temporal Architectural Decisions using Gates.}
%\textcolor{red}{(changed formatting, because "bold" is already used for subsubsection)}
This step takes place after the synapse mechanism, where a specialized neuron $n_a$ is activated.
The specialized neuron's gate parameters $g_a$ consists of $N_{in}$ elements, with each element corresponding to one input channel.
Each gate parameter determines if the corresponding channel is processed using the spatial or temporal operator.

During the forward pass, we sample binary decisions from the gate parameters $g_{a}$ using the Improved Semhash method \cite{kaiser2018discrete,kaiser2018fast,chen2019you}, obtaining a binary vector $\textbf{b} \in \{ 0,1 \}^{N_{in}}$.
Improved Semhash allows us to train gate parameters $g_{a}$ in an end-to-end manner. 
We opt for the Improved Semhash instead of the Gumbel-Softmax here as we can use less parameters ($N_{in}$ instead of $2N_{in}$).
% \ke{(bold b to represent vector?)}
We denote
%If 
the $l$-th element of $\textbf{b}$ as $b_l$. If $b_{l} = 0$, then the corresponding input channel $l$ will use the spatial operator. % and
While if $b_{l} = 1$, then the corresponding input  channel $l$ will use the temporal operator. 
More details of Improved Semhash can be found in the Supplementary.

% \textbf{Specialized Spatio-Temporal Processing.} 
\noindent\underline{Specialized Spatio-Temporal Processing.}
After obtaining channel-wise architectural decisions $\textbf{b}$, we can commence with the channel-wise selection of input feature map $X$ to obtain 
features $X_S$ and $X_T$ as follows, which will be used for learning fine-grained spatial and temporal information respectively:
% \vspace{-0.5em}
\begin{align}
    \label{eqn:XS}
    X_S =(\textbf{1} - \textbf{b}) \cdot  X, \\
    \label{eqn:XT}
    X_T =\textbf{b} \cdot  X,    
    % \vspace{-0.1cm}
\end{align}
% \vspace{-0.5em}
where $\textbf{1}$ is a vector of $1$'s of size $N_{in}$, and $\cdot$ refers to multiplication along the channel dimension while treating each element of $\textbf{b}$ and (\textbf{1} - \textbf{b}) as a channel. 
Using $X_S$ and $X_T$, 
spatial and temporal outputs $Z_{S},Z_T$ are obtained using the respective 
% spatial and temporal operators $S_{a}, T_{a}$
spatial and temporal kernels $S_{a}, T_{a}$
within $n_a$:
% \vspace{-0.3em}
\begin{align}
    Z_{S} = S_{a} (X_S), \\
    Z_{T} = T_{a} (X_T),    
\end{align}
% \vspace{-0.5em}
where $Z_S$ denotes features that capture 
% more \ke{why more? or just spatial?} fine-grained spatial 
spatial information of input feature map $X$, while $Z_T$ denotes features that capture temporal information.
% more \ke{why more? or just temporal?} fine-grained temporal
$Z_S$ and $Z_T$ are then fed to a batch normalization and a ReLU activation layer. We denote the two output features as $Z_S'$ and $Z_T'$.
The output feature map $Z$ is obtained by adding $Z_{S}'$ and $Z_T'$:
% \vspace{-0.1cm}
\begin{equation}
    Z =   Z_{S}' +   Z_{T}'
    % \vspace{-0.1cm}
\end{equation}
% For the $j$-th output channel, we select either the output of the spatial operator and the temporal operator.

% \vspace{-0.5em}
% \begin{equation}
%     Z_a =(\textbf{1} - \textbf{b}) \cdot  Z_{S} +   \textbf{b} \cdot   Z_{T},
% \end{equation}
% where $\textbf{1}$ is a vector of $1$'s of size $N_{out}$, and $\cdot$ refers to multiplication along the channel dimension.

Lastly, a $1 \times 1 \times 1$ convolution is applied to $Z$
% (\ke{why?})
% that acts as a spatio-temporal filter \ke{(i am still not very sure how this is spatial-temporal filter...)} capable of encoding spatio-temporal features. \lgg{used to fuse the two sets of features}
% Those spatio-temporal features are then input to the next DSTS layer or classifier.
%that fuses both features (spatial and temporal) 
to fuse both spatial and temporal features. These fused features $Z'$ are then fed to the next DSTS layer or classifier.
% which is then fed to the next DSTS layer or classifier.

% \begin{equation}
%     Z =(\Vec{1} - b) \cdot  Z_{S} +   b \cdot   Z_{T}
% \end{equation}
% Where $\Vec{1}$ is a vector of $1$'s of size $N_{out}$, and $\cdot$ refers to multiplication along the channel dimension.

% \begin{equation}
%     Z =\textbf{1} (b=0) \cdot  Z_{S} +  \textbf{1} (b=1)  \cdot   Z_{T}
% \end{equation}
% Where $\textbf{1}$ is the indicator function, and $\cdot$ refers to multiplication along the channel dimension.

% Spatio-temporal specialization allows specialized neurons to focus on more spatial or temporal fine-grained information. If a specialized neuron $n_i$ is activated on a subset of similar samples with fine-grained spatial differences, $g_i$ will be trained to produce more $0$'s
% in $\textbf{b}$, such that more channels will be processed by the spatial operation $S_i$ to allow $n_i$ to classify this subset of samples better. On the other hand, if $n_i$ is activated on a subset of similar samples with fine-grained temporal differences, $g_i$ will be trained to produce more $1$'s
% in $\textbf{b}$, such that more of the temporal operation $T_i$ is used for better classification performance.

Spatio-temporal specialization allows specialized neurons to focus on either more spatial or temporal fine-grained information. If a specialized neuron $n_i$ is activated on a subset of similar samples with  fine-grained spatial differences, encoding more spatial information 
% (by applying the spatial operator in more channels)
%(by applying the spatial operator in more channels)
by applying the spatial operator on more channels 
will tend to be more effective, and
$g_i$ will be trained to produce more $0$'s
in $\textbf{b}$.
On the other hand, %if the subset of similar samples
if the samples in the subset contain more
fine-grained temporal differences,
% the model will learn to apply the temporal operator $T_i$ to more channels 
the model will learn to apply the temporal operator 
across more channels,
%(by optimizing $g_i$ to produce more $1$'s in $\textbf{b}$), 
by optimizing $g_i$ to produce more $1$'s in $\textbf{b}$
for better fine-grained action recognition. 
It is also possible that the spatial and temporal aspects are equally important to discriminate similar actions. In this case, $g_i$ will be optimized to %cover
handle 
both spatial and temporal fine-grained information.

\subsection{Upstream-Downstream Learning}
\label{sec:metalearning}

% \lgg{To edit for extra clarity. Motivate better}

% Although our DSTS module is end-to-end trainable, 
% our Upstream-Downstream Learning method further improves its generalization capability
% by meta-learning the model parameters involved in making its dynamic decisions,
% which we call \textit{upstream parameters}.
% These upstream parameters, i.e., scoring kernels $m$ and gate parameters $g$, have profound impacts on the training of \textit{downstream parameters}, i.e., spatial and temporal 
% kernels
% $S$ and $T$.
% This is because upstream parameters determine which downstream parameters are used
% and consequently updated. 
% We use meta-learning 
% \cite{finn2017maml,shu2019metaweightnet} to optimize upstream parameters while taking their downstream effects into account,
% leading to the improved learning of downstream parameters that generalizes better to unseen samples.

To further improve the performance of our DSTS module, we design a UDL algorithm that better optimizes the model parameters involved in making dynamic decisions, which we call \textit{upstream parameters}.
These upstream parameters (i.e., scoring kernels $m$ and gate parameters $g$) that make dynamic decisions and \textit{downstream parameters} (i.e., spatial and temporal operators $S$ and $T$) that process input, are jointly trained during our end-to-end training, 
which can be challenging as \textit{upstream parameters themselves also affect the training of downstream ones}. 
% Specifically, upstream parameters themselves also determine which downstream parameters are used
% and consequently updated.
This is because, upstream parameters determine which downstream parameters will be used, and consequently updated.
Hence, we use meta-learning 
\cite{finn2017maml,shu2019metaweightnet} to optimize upstream parameters while taking their downstream effects into account, leading to the improved learning of downstream parameters and overall improved performance.

% There are three steps in our meta-learning algorithm. In the first step, we simulate an update step by updating downstream parameters while keeping upstream parameters frozen. 
% This simulates the training process of the downstream parameters when the current set of upstream parameters are used to make dynamic decisions.
% In the crucial second step, we evaluate the model's performance on held-out samples in a validation set, measuring its ability to generalize to unseen samples.
% The second-order gradients (with respect to upstream parameters) from this evaluation provide feedback on how upstream parameters can be updated such that their dynamic decisions during training can improve the learning process of downstream parameters, leading to better performance.
% % In the final step, downstream parameters are optimized using the meta-optimized upstream parameters. 
% % These updated downstream parameters are able to generalize better to unseen samples, as compared to after the simulated update.
% In the final step, downstream parameters are optimized using the meta-optimized upstream parameters, which now make dynamic decisions in the model such that downstream parameters are able to benefit more from training and generalize better to unseen samples.
% % In the final step, downstream parameters are optimized using samples from the dynamic decision of meta-optimized upstream parameters, which allow for better training and generalization to unseen samples.

There are three steps in our meta-learning algorithm. In the first step, we simulate an update step by updating downstream parameters while keeping upstream parameters frozen. 
This simulates the training process of the downstream parameters when the current set of upstream parameters are used to make dynamic decisions.
% \textcolor{red}{In the crucial second step, we evaluate the model's performance on held-out samples in a validation set.
% The second-order gradients (with respect to upstream parameters) from this evaluation provide feedback on how upstream parameters can be updated such that their \textit{dynamic decisions during training can improve the learning process of downstream parameters}, leading to better performance. }
In the crucial second step, we evaluate the model's performance on held-out samples in a validation set, which estimates model performance on unseen samples.
% which estimates model performance on unseen samples.
% which measures its improvements on unseen samples.
% measuring its ability to generalize to unseen samples.
% measuring its performance on unseen samples 
% which is an estimate of testing performance.
The second-order gradients (with respect to upstream parameters) from this evaluation provide feedback on how upstream parameters can be updated such that their \textit{dynamic decisions during training can improve the learning process of downstream parameters}, leading to better performance on unseen samples. 
In the final step, downstream parameters are optimized using the \textit{meta-optimized upstream parameters}, which now make dynamic decisions in the model such that downstream parameters are able to benefit more from training and have improved (testing) performance.
% \lgg{to edit}
% \lgg{explain the relevance of unseen samples. e.g. To get evaluations that approximate the performance on unseen samples, the two mini-batches do not contain overlapping samples.}
% \textcolor{red}{}

More concretely, in each iteration, 
we sample two mini-batches from the training data: 
% the input mini-batch is divided into two parts:
% $80\%$ of training samples $D_{train}$ and $20\%$ of validation samples $D_{val}$. 
training samples $D_{train}$ and validation samples $D_{val}$. 
The two mini-batches should not contain overlapping samples, as we want to use $D_{val}$ to estimate performance on unseen samples.
% \textcolor{red}{The two mini-batches should not contain overlapping samples, as we want to evaluate performance on unseen samples.}
% \textcolor{red}{To get evaluations that approximate the performance on unseen samples, the two mini-batches do not contain overlapping samples.}
% \textcolor{red}{To get evaluations that approximate the performance on unseen samples, the two mini-batches do not contain overlapping samples.}
% \textcolor{red}{To get accurate feedback regarding the test-time performance as best as we can, the two mini-batches do not contain overlapping samples.}
% The two mini-batches should not contain overlapping samples, as we want to evaluate the test-time performance as best as we can.
% As the purpose of the validation samples is to evaluate the generalizability of the simulated update, the two mini-batches should not contain overlapping samples. %samples should not overlap between the two mini-batches.
The algorithm proceeds in three steps:

Firstly, a \textbf{Simulated Update Step} updates downstream parameters $d$ 
% (consisting of spatial operators $S$ and temporal operators $T$)
using supervised loss $\ell$ on $D_{train}$.
%  \vspace{-0.2cm}
\begin{equation}
    \label{eqn:simulated_downstream_update}
    \hat{d} = d- \alpha  \nabla_{d} \ell (u,d;D_{train}),
\end{equation}
% \vspace{-0.5em}
where $\alpha$ is a learning rate hyperparameter, while $u$ and $d$ denote the upstream and downstream parameters respectively. We keep upstream parameters $u$ fixed in this step.

Secondly, a \textbf{Meta-Update Step} evaluates the updated model on $D_{val}$. We 
update upstream parameters $u$ using the second-order gradients with respect to $u$ 
% (when they were used to make decisions in the first Simulated Update Step) as follows:
when they were used to make decisions in the first Simulated Update Step, as follows:
% used in the first step.
% \vspace{-0.3cm}
\begin{equation}
    % u' = u- \alpha  \nabla_{u} \ell (d- \alpha  \nabla_{d} \ell (d,u;D_{train}),\hat{u};D_{val})
    \label{eqn:metaupdate}
    u' = u- \alpha  \nabla_{u} \ell  (\hat{u},\hat{d};D_{val}),
    % \vspace{-0.5em}
\end{equation}
% \vspace{-0.5em}
% \ke{in skel paper, you also do not just put $\hat{d}$,  then the \hat{d} at first step does not make sense at all? or maybe the first step aims  to just  give you the gradient? or now $d=\hat{d}$? the algorithm is much more clear than here. maybe try to adjust based on the algorithm}
% where $\hat{u} = u$,
% (\ke{weird, why not just u?} \lgg{because the second-order gradients are computed with respect to the $u$ that is used in the first step in Eq.8. So using $u$ anywhere else might make it somewhat imprecise and incorrect })
% where $\hat{u} = u$,
where $\hat{u}$ is a copy of $u$,
but no gradients are computed with respect to $\hat{u}$. 
We denote it this way, because the same set of $u$ parameters are used twice (in Eq.~\ref{eqn:simulated_downstream_update} and Eq.~\ref{eqn:metaupdate}), and we want to compute second-order gradients $\nabla_u$ with respect to $u$ in Eq.~\ref{eqn:simulated_downstream_update}, not first-order gradients with respect to $\hat{u}$ in Eq.~\ref{eqn:metaupdate}.
% Roughly, we want to tune the $u$ from the first step, to optimize the validation loss in the 2nd step.
% \ke{sorry, still confuse, u=uhat, why not just u, as u' is the updated but u is still the same as previous equation, also algorithm does not mention uhat}
% \lgg{Because the same set of upstream parameters $u$ are actually used twice -- once in the first step, and once in the second. We want to get second-order gradients with respect to the $u$ in the first step, not the second usage (which I currently denote as $\hat{u}$). It seems a bit confusing to me if we use $u$ for both usages, especially since the gradient $\nabla_u$ takes gradients with respect to $u$, not $\hat{u}$ }\ke{but $\hat{u}$ is u, the u in second step is u'. i think maybe ask Jun to check. i may be wrong.} \lgg{u' in the second step is the updated result (from u). The set of upstream parameters used in the second step is indeed u (or $\hat{u}$), but we are not tuning this u based on their performance in this step. $>$.$<$}
% Note that second order gradients $\nabla_u$ are taken with respect to $u$  in Eq.~\ref{eqn:simulated_downstream_update}.
These second-order gradients $\nabla_u$ provide feedback on how to adjust $u$ such that their dynamic decisions lead to better training of the downstream parameters (as simulated in the Simulated Update Step), resulting in improved performance on unseen samples.
$d$ is not updated in this step. 

% with additional training, improves performance on unseen samples.
% The second-order gradients provide feedback on how to adjust $u$ in such a way that, with additional training, improves performance on unseen samples.
% The meta-gradient updates gate parameters such that the model shows improved performance when additional updates of the operators are considered.

Finally, $d$ is updated in the \textbf{Actual Update Step} while keeping $u'$ frozen. 
\begin{equation}
    d' = d- \alpha  \nabla_{d} \ell (u',d;D_{train})
\end{equation}
One iteration of this algorithm concludes here, and we obtain updated parameters $u'$ and $d'$. 
An outline of the algorithm is shown in the Supplementary. 
\section{Experiments}

%\subsection{Datasets}

We conduct experiments using our proposed DSTS module on two popular fine-grained action recognition datasets, % We experiment on two popular datasets, 
i.e., the Something-Something v2 dataset (SSV2) \cite{goyal2017something} and Diving48 dataset \cite{li2018diving48}. 

\textbf{SSV2} \cite{goyal2017something} 
is a large dataset, containing approximately 220k videos across 174 different classes. It consists of crowd-sourced clips that show humans performing basic actions with various types of everyday objects. The difference between classes could lie in fine-grained spatial or temporal details, as depicted in Fig.~\ref{fig:examples}. Following \cite{goyal2017something,liu2021video,zhang2021morphmlp}, we split the data into 169k training and 27k test videos. %
%We do not use the 24k validation samples.
%169k training/24k validation/27k testing

\textbf{Diving48} \cite{li2018diving48} contains approximately 18k trimmed video clips of 48 classes of competitive diving sequences. 
There are fine-grained differences between the 48 %dive sequences 
classes, which could exist at takeoff, in flight, at entry, or a combination of them in the  diving sequences, making it a challenging classification task. %that coarse-grained action recognition methods are unable to tackle successfully.\ke{(just to avoid providing evidence or reference.)}
Following \cite{li2018diving48,zhang2021tqn}, we split the data into 16k training and 2k test videos. 
% The video lengths have a very wide range: min: 24 frames, max: 822 frames, average: 158 frames. 
Following \cite{zhang2021tqn}, we use the cleaned (v2) labels released in Oct 2020.

% \vspace{-0.2cm}
\subsection{Implementation details}
% \vspace{-0.1cm}
% \noindent\textbf{Network Architecture.} 
To evaluate the efficacy of the proposed DSTS module, Swin-B transformer \cite{liu2021video} and TPN \cite{yang2020tpn} are used as the backbone networks. In our experiments, each DSTS layer contains 10 specialized neurons ($N=10$) and the DSTS module has 3 layers ($L=3$). 
The dimensions of the input $X$, such as $N_{in},N_{t},N_{h},N_{w}$ are determined by different backbone networks, and we set $N_{out}=N_{in}$. 
% and thus $N_{out}$ \ke{isnt $N_{out}$ decided manually?}, $N_{in}$ 
Thus, the shape of
$S_{ij},T_{ij},g_{ij}$ and $m_{ij}$ for each $n_{ij}$ % $b_{ij}$ 
are dependent on the backbones. 
% The $1 \times 1 \times 1$ convolution kernel has the shape of $N_{out} \times N_{out} \times 1 \times 1 \times 1$.
% The spatial operator $S_{ij}$ is a block consisting of a $1 \times 3 \times 3$ 2D convolutional layer, a batch norm layer and an ReLU activation function, and the temporal operator $T_{ij}$ is a block consisting of a $3 \times 1 \times 1$ 1D convolutional layer, a batch norm layer and an ReLU activation function. 
% The spatial operator $S_{ij}$ is a block consisting of a $1 \times 3 \times 3$ 2D convolution,
% a batch norm and a ReLU, and the temporal operator $T_{ij}$ is a block consisting of a $3 \times 1 \times 1$ 1D convolution, a batch norm and a ReLU.
% \ke{in the method, the operator just mention conv kernel size,  in the method after eq7 i have mention the two features are then fed to a batch normalization and a RELU before addition. if suitable can delete the previous two.}
% The scoring convolution $m_{ij}$ comprises of a $1 \times 1 \times 1$ convolution and a 
%% linear layer to downsample the input feature $X$
% summation to obtain a scalar impulse $v_{ij}$.
% \ke{in the method, these two steps are separated, i think no need to mention the scoring here like this.}

% \noindent\textbf{Training.} 
% For \textbf{training}, 
The experiments are conducted on 8 Nvidia V100 GPUs with batch size $B=8$. We follow the experimental settings of Video Swin Transformer \cite{liu2021video}, using the AdamW optimizer and setting the initial learning rate $\alpha$ as $3\times10^{-4}$. For TPN, we follow the experimental settings in \cite{yang2020tpn}, using the SGD optimizer and setting the initial learning rate $\alpha$ as $0.01$. 
We compute cross-entropy loss as the supervised loss $\ell$ %from
%in the classification task.
for classification.
During \textbf{training}, using the Gumbel-Softmax and Improved Semhash techniques for selection of specialized neurons and operators, our model is end-to-end trainable.
% for each layer $j$, its input feature $X$ 
% is fed into all specialized neurons $\{n_{ij} \}_{i=1}^N$ to obtain impulse values $\{v_{ij}\}_{i=1}^N$ and outputs $\{Z'_{ij}\}_{i=1}^N$ from all $N$ of them.
% In each specialized neuron $n_{ij}$, gate parameters $g_{ij}$ are input into the Improved Semhash method to obtain the binary vector $\textbf{b}$. The obtained
% $\textbf{b}$ and $(\textbf{1}-\textbf{b})$ are multiplied to the input $X$ as shown in Eq.~\ref{eqn:XS} and Eq.~\ref{eqn:XT}, and those outputs 
% are then processed by the spatial and temporal operator respectively, to eventually obtain the specialized neuron output $Z'_{ij}$.
% Thus, gates parameters $g_{ij}$ are end-to-end trainable using gradients flowing through $Z'_{ij}$. 
% Then, the Gumbel-Softmax technique is used to activate a specialized neuron $n_{aj}$, taking in $\{v_{ij}\}_{i=1}^N$ and returning a one-hot vector of length $N$ with a $1$ at index $a$.
% This one-hot vector is multiplied to the outputs $\{Z'_{ij}\}_{i=1}^N$ of all $N$ specialized neurons, effectively masking out the specialized neurons that are not activated, while allowing for all scoring kernels $\{m_{ij} \}_{i=1}^N$ to be trained end-to-end. 
We set Gumbel-Softmax temperature $\tau = 1$, and the noise applied to Improved SemHash is sampled from a standard Gaussian distribution.
% , i.e., the mean equals $0$ and the variance equals $1$. 

During \textbf{testing}, given an input feature map $X$, impulse values $\{v_{ij}\}_{i=1}^N$ are computed for all $N$ specialized neurons in each layer $j$. However, because we do not require gradients this time, 
the input $X$ is only processed 
by the best-matching specialized neuron $n_{aj}$ of each layer $j$, to obtain output $Z'_{aj}$.
% only computation of the best-matching specialized neuron $n_{aj}$ is required, to obtain output $Z_a$.
% Within $n_{aj}$, gates $g_{aj}$ are input into Improved Semhash to obtain $\textbf{b}$ for channel-wise architecture selection as usual.
Notably, no noise is added to Gumbel-Softmax and Improved Semhash during inference.

\subsection{Experiment Results}
% \vspace{-0.1cm}

\noindent{\textbf{Results on SSv2.}}
Following \cite{liu2021video,arnab2021vivit,lin2019tsm}, we report Top-1 and Top-5 accuracy scores across all models on the test set of SSv2.
Results are shown in Table \ref{tab:ssv2}. 
% There was a Top-1 improvement of 2.5\% on the CNN-based TPN and 2.2\% on the Transformer-based Swin. Swin w/ DSTS achieved a Top-1 accuracy of 71.8\%, surpassing the current state-of-the-art \cite{liu2021video}. 
As both CNNs and Transformers are used to tackle action recognition, we test DSTS on a CNN-based architecture (TPN~\cite{yang2020tpn}) and a Transformer-based architecture (Swin-B~\cite{liu2021video}) to investigate if our DSTS module provides performance gains on both types of architectures.

Adding our DSTS module to baseline architectures leads to improved performance on both architectures. Adding DSTS to TPN (\textbf{TPN w/ DSTS}),
the performance of TPN improves by 2.5\%, achieving a Top-1 accuracy of 67.2\%. To the best of our knowledge,
this performance is state-of-the-art
% this is state-of-the-art performance 
among CNN-based architectures, surpassing even the performance of the two-stream TSM which utilizes additional optical flow information. This shows that DSTS can improve performance for CNN-based backbones on fine-grained action recognition.
Adding DSTS to Swin-B (\textbf{Swin-B w/ DSTS}) improves Top-1 accuracy by 2.2\%, achieving a new state-of-the-art of 71.8\%, showing that DSTS can help improve fine-grained action recognition on Transformer-based backbones as well. Qualitative results and visualizations have been placed in the Supplementary.

\begin{table}[tb]
\scriptsize
\caption{Top-1 and Top-5 scores (\%)  on SSv2. Type ``C" indicates CNN-based architectures and ``T" indicates Transformer-based architectures.
Our DSTS module improves Top-1 accuracy of TPN by 2.5\% and Swin-B by 2.2\%. 
% \lgg{Should we merge both tables?}
% \vspace{-0.3cm}
}
\centering
\begin{tabular}{|l |c| c| c|} 
% \toprule
\hline
Method & Type & Top-1 & Top-5 \\
% \midrule 
\hline
\hline
SlowFast \cite{feichtenhofer2019slowfast} & C & 63.1 & 87.6 \\
TPN \cite{yang2020tpn} & C & 64.7 & 88.1 \\
% TimeSformer \cite{bertasius2021space} & 62.5 & \\
ViViT-L \cite{arnab2021vivit} & T & 65.4 & 89.8 \\
%% TSM (RGB) \cite{lin2019tsm} & 63.3 \\
%%taken straight from TSM paper
TSM (Two-stream) \cite{lin2019tsm} & C & 66.6 & 91.3 \\
MViT-B \cite{fan2021multiscale} & T & 67.7 & 90.9\\
Swin-B \cite{liu2021video} & T & 69.6 & 92.7\\
% \midrule
\hline
\hline
TPN w/ DSTS & C & 67.2 & 89.2 \\
Swin-B w/ DSTS & T & \textbf{71.8} & \textbf{93.7}\\
% \bottomrule
\hline
\end{tabular}
\label{tab:ssv2} 
% \vspace{-0.2cm}
\end{table}

%VIMPAC: Video Pre-Training via Masked Token Prediction and Contrastive Learning
%OBJECT-REGION VIDEO TRANSFORMERS

% \lgg{the papers generally report mean acc over classes as well? (to prevent overfitting to the largest classes)}
\noindent{\textbf{Results on Diving48.}}
Following \cite{zhang2021tqn}, we report Top-1 accuracy and mean accuracy per class across all models on Diving48 dataset. Results are shown in Table \ref{tab:diving48}.
Using DSTS module leads to significant improvements on Diving48 as well. It achieves a Top-1 improvement of 2.2\% on TPN and 2.5\% on Swin-B.
TPN w/ DSTS achieves state-of-the-art result of 88.4\% Top-1 accuracy.
% \lgg{There was an improvement of \% on the CNN-based TPN and \% on the Transformer-based Swin. Swin w/ DSTS achieved a Top-1 accuracy of \%, surpassing the current state-of-the-art by \%. }

% %most results taken from tqn paper
% \begin{table}[tb]
% \footnotesize
% \caption{Top-1 accuracy scores on Diving48.}
% \centering
% \begin{tabular}{l c } 
% \toprule
% Method & Top-1 \\
% \midrule 
% I3D \cite{carreira2017i3d} & 48.3 \\
% TSM (Two-stream) \cite{lin2019tsm} & 52.5 \\
% %  \cite{wu2019longterm}
% GST \cite{luo2019grouped} & 78.9 \\
% TQN \cite{zhang2021tqn} & 81.8 \\
% TPN \cite{yang2020tpn} & \\
% Swin-B \cite{liu2021video} &  \\
% \midrule
% TPN w/ DSTS &    \\
% Swin w/ DSTS &  \\
% \bottomrule
% \end{tabular}
% \label{tab:diving48} 
% \end{table}

%most results taken from tqn paper
\begin{table}[tb]
\scriptsize
\caption{Top-1 and Class-wise accuracy scores (\%)  on Diving48. 
% Type ``C" indicates CNN-based architectures and ``T" indicates Transformer-based architectures.
Our DSTS module improves Top-1 accuracy of TPN by 2.2\% and Swin-B by 2.5\%.
% \vspace{-0.3cm}
% \lgg{Results are taken from tqn paper which reproduced everything using the cleaned labels }
}
\centering
\begin{tabular}{|l |c |c |c|} 
% \toprule
\hline
Method & Type & Top-1 & Class-wise Acc \\
% \midrule 
\hline
\hline
I3D \cite{carreira2017i3d} & C & 48.3 & 33.2 \\
TSM (Two-stream) \cite{lin2019tsm} & C & 52.5 & 32.7 \\
%  \cite{wu2019longterm}
GST \cite{luo2019grouped} & C & 78.9 & 69.5 \\
TQN \cite{zhang2021tqn} & T & 81.8 & 74.5 \\
Swin-B \cite{liu2021video} & T & 80.5 & 69.7 \\
TPN \cite{yang2020tpn} & C & 86.2 & 76.0  \\
% \midrule
\hline
\hline
Swin-B w/ DSTS & T & 83.0 & 71.5  \\
TPN w/ DSTS & C & \textbf{88.4} &  \textbf{78.2} \\
% \bottomrule
\hline
\end{tabular}
\label{tab:diving48} 
% \vspace{-0.2cm}
\end{table}

% \noindent{\textbf{Qualitative observations.}}

% \vspace{-0.1cm}
\subsection{Ablation Studies}
\label{sec:ablation}
% \vspace{-0.1cm}
We conduct extensive ablation studies to evaluate the importance of certain design choices. Ablation studies are conducted on Diving48, using TPN as a backbone. More experiments are placed in the Supplementary.

\begin{wraptable}{r}{0.44\linewidth}
\scriptsize
\caption{Evaluation results (\%) on
the impact of  
 spatio-temporal specialization of DSTS modules on Diving48. 
%  \vspace{-0.2cm}
}
\centering
\begin{tabular}{|l |c| c|} 
% \toprule
\hline
Method  & Top-1 & Class-wise Acc \\
% \midrule
\hline
\hline
DSTS w/o STS & 87.2 & 76.5  \\
DSTS w/o Gates & 87.3 & 76.7 \\
DSTS w/ STS & 88.4 & 78.2  \\
% \bottomrule
\hline
\end{tabular}
\label{tab:ablation_STS}
% \vspace{-0.2cm}
\end{wraptable}

% \noindent{\textbf{1) Spatio-temporal specialization.}}
\noindent{\textit{1) Spatio-temporal specialization.}}
We evaluate the impact of spatio-temporal specialization on our DSTS module, and the results are shown in Table \ref{tab:ablation_STS}.
It can be observed that DSTS module with spatio-temporal specialization (\textbf{DSTS w/ STS}) performs better than DSTS without it (\textbf{DSTS w/o STS}), showing its effectiveness. 
% showing the effectiveness of spatio-temporal specialization.
For DSTS w/o STS, only one operator, i.e., a 3D convolution with batch normalization and ReLU, within each $n_{ij}$ is employed to process $X$. 
Besides, when we remove gates $g_{ij}$ (\textbf{DSTS w/o Gates}), and let all specialized neurons have the same factorized architecture (the channels are split into two fixed halves, 
% to which are applied the spatial and temporal operators respectively), 
to which the spatial and temporal operators are applied respectively), 
%half spatial and half temporal
% \ke{(randomly split or each input channel randomly applied a spatial/temporal)?}),
the performance decreases by 1.1\%. This shows that our gates learn channel-wise architectures that are more specialized and effective for fine-grained recognition, compared to fixed architectures.

\begin{wraptable}{r}{0.62\linewidth}
\scriptsize
\caption{Evaluation results (\%)  on
% Performance comparison (\%) of 
the impact of the synapse mechanism on Diving48. 
% \textcolor{cyan}{The model size is too large. Perhaps do not put in a table?} 
%For DSTS w/o synapse mechanism setting, we average the output of all specialized neurons in each layer.
% different number of activated specialized neurons on Diving48. 
% \lgg{To add baseline model + column for parameter count}
%  \vspace{-0.2cm}
}
\centering
\begin{tabular}{|l| c| c| c|} 
% \toprule
\hline
Method & Top-1 & Class-wise Acc & Model Size \\
% \midrule
\hline
\hline
Baseline TPN & 86.2 & 76.0 & 63M \\
w/o Synapse Mechanism & 86.5 & 76.4 & 75M \\
w/ Synapse Mechanism & 88.4 & 78.2 & 75M  \\
% \bottomrule
\hline
\end{tabular}
\label{tab:ablation_selected_number}
% \vspace{-0.1cm}
\end{wraptable}

% \noindent{\textbf{2) Number of activated specialized neurons.} }
% \noindent{\textit{2) Number of activated specialized neurons.}} 
\noindent{\textit{2) Synapse Mechanism.}}
% We investigate whether activating more specialized neurons would lead to better performance, and results are shown in Table \ref{tab:ablation_selected_number}.
We investigate the impact of the synapse mechanism, and results are shown in Table \ref{tab:ablation_selected_number}.
Following our method with the dynamic synapse mechanism and activating the most relevant specialized neuron at each layer (\textbf{w/ Synapse Mechanism}) achieves better results compared to activating all specialized neurons and averaging their outputs (\textbf{w/o Synapse Mechanism}), which is a non-dynamic design with the same number of parameters as our method.
This shows that the performance improvement comes from our synapse mechanism and its dynamic design, and not the additional parameters.
This improvement is because, unlike w/o Synapse Mechanism which trains all neurons on all data samples and tends to learn more common %general
discriminative cues that apply to the more common samples, our DSTS module trains each specialized neuron only on a subset of similar samples, explicitly pushing them to gain better specialized fine-grained abilities.
% \textcolor{red}{We also observe that, our model with the dynamic Synapse Mechanism outperforms a variant model with the same number of parameters, but without the dynamic design, showing performance improvement comes from our dynamic design, and not additional parameters.}
% Our dynamic design achieves a significant improvement over a non-dynamic variant with the same number of parameters, showing that the performance improvement comes from our dynamic design and not the additional parameters.
% Our dynamic design shows clear improvement over non-dynamic variant with \textbf{same number of parameters}, showing performance improvement comes from our dynamic design, and not additional parameters.

% \noindent{\textbf{3) Upstream-Downstream Learning.}}
\noindent{\textit{3) Upstream-Downstream Learning.}}
We conduct experiments to evaluate the performance gains from our UDL method, and the results can be seen in Table \ref{tab:ablation_meta}. 
\begin{wraptable}{r}{0.48\linewidth}
% \vspace{-0.2cm}
\scriptsize
\caption{Evaluation results (\%)  on
% Performance comparison (\%) of 
the impact of
the UDL method on Diving48. 
%  \vspace{-0.2cm}
}
\centering
\begin{tabular}{|l |c| c|} 
% \toprule
\hline
Method  & Top-1 & Class-wise Acc \\
% \midrule 
\hline
\hline
DSTS w/o UDL & 87.4 & 76.7  \\
DSTS w/ UDL & 88.4 & 78.2  \\
% \bottomrule
\hline
\end{tabular}
\label{tab:ablation_meta}
% \vspace{-0.2cm}
\end{wraptable}
We observe that our UDL method (\textbf{DSTS w/ UDL}) improves performance over using backpropagation in a single step (\textbf{DSTS w/o UDL}). 
We emphasize that such these performance gains are achieved using only slightly more training time, which is reported in the Supplementary.

\begin{wraptable}{r}{0.34\linewidth}
% \vspace{-0.3cm}
\scriptsize
\caption{Evaluation results (\%)  for
% Performance comparison (\%) of 
different numbers
of specialized neurons $N$ in each DSTS module on Diving48. 
%  \vspace{-0.2cm}
}
\centering
\begin{tabular}{|l |c| c| } 
% \toprule
\hline
$N$  & Top-1  & Class-wise Acc \\
% \midrule 
\hline
\hline
5 & 87.3 & 76.2  \\
10 & 88.4 & 78.2  \\
15 & 88.3  & 78.2 \\
% \bottomrule
\hline
\end{tabular}
\label{tab:ablation_specialized_neurons}
% \vspace{-0.2cm}
\end{wraptable}

% \textbf{Improved Semhash parameter}

% \noindent{\textbf{4) Number of specialized neurons in each DSTS layer.}} 
\noindent{\textit{4) Number of specialized neurons in each DSTS layer.}} 
We evaluate the impact of using different numbers of specialized neurons in each DSTS layer,
and the results are shown in Table \ref{tab:ablation_specialized_neurons}. 
When $N$ is low (e.g., $N=5$), using more specialized neurons in each DSTS layer (e.g., $N=10$) improves the performance, which can be explained by the increase in representational capacity. More precisely, when there are more specialized neurons, each one can afford to be more specialized towards a smaller subset of data, which improves their capability. We use $N=10$ as this improvement effect tapers off when $N$ is increased beyond $10$.
% , the improvement yielded from adding extra specialized neurons is low, at \lgg{??}. Thus, we decided to set the number of specialized neurons, $N = ?$.

%%%no wrap
% \begin{table}[tb]
% \footnotesize
% \caption{Evaluation results (\%)  for
% % Performance comparison (\%) of 
% different numbers
% of specialized neurons $N$ in each DSTS module on Diving48. 
%  \vspace{-0.2cm}
% }
% \centering
% \begin{tabular}{|l |c| c| } 
% % \toprule
% \hline
% $N$  & Top-1  & Class-wise Acc \\
% % \midrule 
% \hline
% \hline
% 5 & 87.3 & 76.2  \\
% 10 & \textbf{88.4} & \textbf{78.2}  \\
% 15 & 88.3  & 78.1 \\
% % \bottomrule
% \hline
% \end{tabular}
% \label{tab:ablation_specialized_neurons}
% \vspace{-0.2cm}
% \end{table}

%%%%%%%%%%%%%%  deleted from camera ready %%%%%%%%
\begin{wraptable}{r}{0.34\linewidth}
% \vspace{-0.3cm}
\scriptsize
\caption{Evaluation results (\%)  for
% Performance comparison (\%) of 
different numbers %number
of DSTS layers $L$ on Diving48. 
%  \vspace{-0.2cm}
}
\centering
\begin{tabular}{|l |c| c|} 
% \toprule
\hline
$L$ & Top-1 & Class-wise Acc \\
% \midrule 
\hline
\hline
1 & 87.5 & 76.8 \\
3 & 88.4 & 78.2 \\
5 & 88.2 & 78.2 \\
% \bottomrule
\hline
\end{tabular}
\label{tab:ablation_layers}
% \vspace{-0.2cm}
\end{wraptable}

\noindent{\textit{5) Number of DSTS layers.}}
We also evaluate the impact of varying $L$, i.e., stacking different numbers of DSTS layers, and results are shown in Table \ref{tab:ablation_layers}. 
As expected, stacking more DSTS layers leads to better performance. This is because, the DSTS module with more layers could have greater representational capacity to process more complex and fine-grained cues. When we increase $L$ from $1$ to $3$, we obtain an improvement of $0.9 \%$, and increasing $L$ further does not lead to further improvement. We thus set $L=3$.

%%not wrapped
% \begin{table}[tb]
% \footnotesize
% \caption{Evaluation results (\%)  for
% % Performance comparison (\%) of 
% different numbers %number
% of DSTS layers $L$ on Diving48. 
%  \vspace{-0.2cm}
% }
% \centering
% \begin{tabular}{|l |c| c|} 
% % \toprule
% \hline
% $L$ & Top-1 & Class-wise Acc \\
% % \midrule 
% \hline
% \hline
% 1 & 87.5 & 76.8 \\
% 3 & \textbf{88.4} & \textbf{78.2} \\
% 5 & 88.1 & 78.0 \\
% % \bottomrule
% \hline
% \end{tabular}
% \label{tab:ablation_layers}
% \vspace{-0.2cm}
% \end{table}

% \vspace{-0.2cm}
\section{Conclusion}
In this paper, we have proposed a novel DSTS module consisting of dynamically activated specialized neurons for fine-grained action recognition. 
% Our spatio-temporal specialization method further optimizes the architectures of specialized neurons to focus more on spatial or temporal aspects.
Our spatio-temporal specialization method optimizes the architectures of specialized neurons to focus more on spatial or temporal aspects.
% An Upstream-Downstream Learning procedure improves the generalization capability of our DSTS module.
Our UDL procedure further improves the performance of our DSTS module. 
We obtain state-of-the-art fine-grained action recognition performance on two popular datasets by adding DSTS modules to baseline architectures.

% \footnotesize \noindent \textbf{Acknowledgement} This work is supported by National Research Foundation, Singapore under its AI Singapore Programme (AISG Award No: AISG-100E-2020-065), SUTD Startup Research Grant and MOE Tier 1 Grant. This work is also partially supported by Natural Science Foundation of China (NSFC) under the Grant no. 62172285

\footnotesize \noindent \textbf{Acknowledgement}
This work is supported by National Research Foundation, Singapore under its AI Singapore Programme (AISG Award No: AISG-100E-2020-065), Ministry of Education Tier 1 Grant and SUTD Startup Research Grant. This work is also partially supported by Natural Science Foundation of China (NSFC) under the Grant no. 62172285. The research is also supported by TAILOR, a project funded by EU Horizon 2020 research and innovation programme under GA No 952215.

\clearpage
\bibliographystyle{splncs04}
\bibliography{egbib}
\end{document}